\def\year{2022}\relax
\documentclass[letterpaper]{article} 
\usepackage{booktabs}
\usepackage{amsmath} 
\usepackage{algorithm}
\usepackage{algpseudocode} 
\usepackage{xcolor}

\usepackage{aaai22}  
\usepackage{times}  
\usepackage{helvet} 
\usepackage{courier}  
\usepackage[hyphens]{url}  
\usepackage{graphicx} 
\usepackage{natbib}  
\usepackage{caption} 
\usepackage{color}

\nocopyright

\title{Cyberbullying detection across social media platforms via platform-aware adversarial encoding}
\author{
   Peiling Yi,
   Arkaitz Zubiaga
\\
}
\affiliations{
    Queen Mary University of London, UK\\

   \{p.yi, a.zubiaga\}@qmul.ac.uk

}

\begin{document}

\maketitle

\begin{abstract}
Despite the increasing interest in cyberbullying detection, existing efforts have largely been limited to experiments on a single platform and their generalisability across different social media platforms has received less attention. We propose XP-CB, a novel cross-platform framework based on Transformers and adversarial learning. XP-CB can enhance a Transformer leveraging unlabelled data from the source and target platforms to come up with a common representation while preventing platform-specific training. To validate our proposed framework, we experiment on cyberbullying datasets from three different platforms through six cross-platform configurations, showing its effectiveness with both BERT and RoBERTa as the underlying Transformer models.
\end{abstract}

\section{Introduction}

Cyberbullying is a form of bullying that is perpetrated through online devices \cite{Smith2008}. With the growth in usage of digital devices and Internet platforms such as social media, cyberbullying has become a major problem worldwide \cite{Nixon2014}. This has motivated research in cyberbullying detection as the predictive task aiming to identify cyberbullying posts for enabling harm prevention \cite{Rosa2019}.

Despite increasing efforts in furthering research in cyberbullying detection, existing methods have been predominantly investigated in a single social media platform. There is however increasing evidence that classifiers built for and tested on a particular social media platform tend to underperform when applied to new platforms \cite{yin2021towards}, limiting their generalisability. Generalisability of models to other platforms has been barely studied, not least in the context of cyberbullying detection \cite{mladenovic2021cyber}. Recent models for contextualised embeddings based on Transformer models, such as BERT \cite{Devlin2019} and RoBERTa \cite{liu2019roberta}, are promising alternatives that can provide some generalisability through their ability to transfer knowledge. Still, they have been shown to struggle in situations where there is a big drift from source to target data \cite{Sun2019a}.

To increase the potential of Transformer models when applied to a different social media platform, we propose XP-CB, a novel platform-aware adversarial framework for cross-platform cyberbullying detection. By training a classifier on a source platform, for which labelled data is available, we aim to test its ability to generalise to a new target platform, for which labelled data is lacking. To mitigate the effects of platform data shift, the core intuition of our proposed framework is to combine a multi-Transformer embedding alignment strategy with an adversarial network to reconstruct the target Transformer encoder. The target Transformer encoder is forced to map the target input to the source Transformer latent representation space, achieving more similar content representations for both source and target platforms. The classifier trained on the source platform data can be subsequently applied to the target platform. The contributions of this paper can be summarised as follows:

\begin{itemize}
 \item We propose XP-CB, which is to the best of our knowledge the first framework for cross-platform cyberbullying detection, operationalised by furthering the potential of a Transformer model through the integration of an adversarial network.
 \item To assess the potential of XP-CB, we perform cyberbullying detection experiments in six cross-platform configurations involving three datasets from Formspring, Twitter and Wikipedia. These platforms present very different characteristics, particularly when it comes to the lenght of the posts.
 \item We show that XP-CB can achieve state-of-the-art performance by consistently outperforming a competitive model as well as vanilla Transformer models.
\end{itemize}

\section{Related work}

\subsection{Cyberbullying detection}

Early methods to cyberbullying detection relied on rule-based methods \cite{Mahmud2008,Nahar2014}, focused on feature engineering \cite{hosseinmardi2015analyzing} and used lexica \cite{Dadvar2012,VanHee2018}. More recent methods use word embeddings along with deep learning models to build  more discriminative models, leading to improved performance \cite{Yuvaraj2021,Cheng2020}. Still, this research predominantly focuses on tackling cyberbullying on a single platform, which limits the potential of transferring existing models to new, unseen social media platforms where labelled data is lacking.

Cross-platform cyberbullying detection is still in its infancy, which was pioneered by \cite{Agrawal2018}. They studied the performance of a zero-shot transfer learning approach on three different social platforms (Wikipedia, Twitter and Formspring), training and testing on different platforms. Their study highlighted the challenging nature of the problem, finding that the three datasets exhibit different forms of cyberbullying with limited feature overlap across platforms. By using a  Bidirectional LSTM (BiLSTM) model coupled with lists of swear words to enable transferability, they still showed a substantial performance drop from running experiments within a platform, to switching to experiments across platforms. To our knowledge, the only other work on cross-platform cyberbullying detection is that by \citet{Dadvar2018}, who further tested the above BiLSTM model on a new platform, YouTube, leading to similar findings and highlighting the need for better models that further generalisability across platforms.

Rather than finding an overlap of features, our aim with XP-CB is to enable Transformer models the capacity of defining a latent feature space that reconciles the differences between the source and target platforms.

\subsection{Adversarial networks}

Adversarial adaptation methods have become increasingly popular for domain adaptation, which seek to minimise the variance between source and target data through an adversarial objective \cite{Tzeng2017a}. These methods are motivated by Generative Adversarial Networks \cite{Creswell2018}, which consist of two parts: (1) the Generator is trained to generate synthetic instances in a way that confuses the discriminator; and (2) the Discriminator, responsible in turn for trying to distinguish the samples created by the generator. In the process of adversarial adaptation across divergent data sources, the roles of synthetic instances and real instances can be replaced with training samples and test samples, i.e. in cross-platform experiments, the role of the discriminator becomes that of distinguishing if an instance belongs to the source or target platform \cite{Creswell2018}.

Inspired by this trend, we propose the integration of a Transformer with an adversarial adaptation component for the cross-platform cyberbullying detection task. We build on ADDA (Adversarial Discriminant Domain Adaptation) \cite{Tzeng2017a} as the adversarial component. ADDA is a general network that enables combining a discriminative model, weight sharing, and a GAN loss for effectively training a robust and adaptive Deep Neural Network (DNN).

\section{XP-CB: Model architecture}
\begin{figure}[h]
  \centering
  \includegraphics[width=\linewidth]{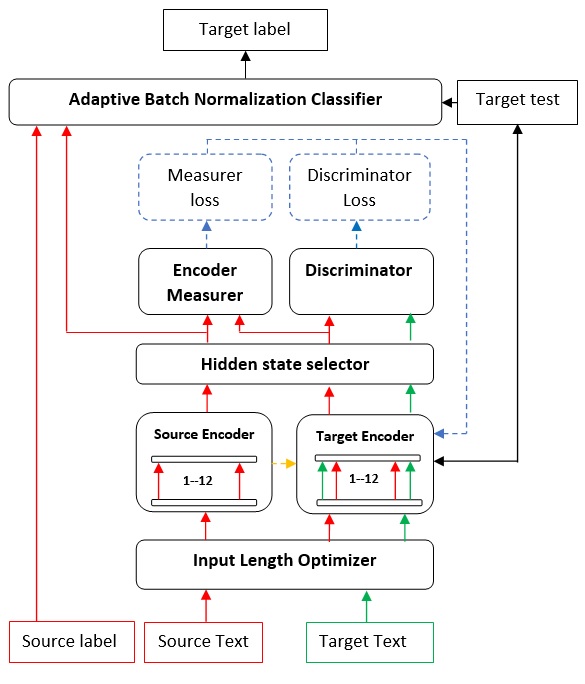}
  \caption{XP-CB Model Architecture. Red: source platform data flow. Green: target platform data flow. Black: test data flow. Blue (dashed): how the loss is fed back to the back propagation algorithm. Orange (dashed): source encoder parameters, used to initialise the target encoder.}
  \label{fig:architecture}
\end{figure}

XP-CB is an end to end framework (see Figure \ref{fig:architecture}), whose components can be trained at different times. The overarching objective of XP-CB is to perform dual alignment, which are operationalised by different components:

\begin{itemize}
 \item \textbf{The embedding alignment} is responsible for injecting different cross-platform fine-tuning strategies into the framework components, which aims to improve the encoder's adaptability to new platforms. It includes three subcomponents: Input Length Optimiser, Hidden States Selector and Adaptive Batch Normalisation classifier.
 \item \textbf{The Adversarial alignment} is responsible for integrating the GAN methodology and the ADDA framework to align the target input representation to the source latent embedding space, which include the Encoder Measurer and Discriminator subcomponents.
\end{itemize}

\textbf{Input Length Optimiser.} To deal with the different lengths across platforms, a possible approach would be to truncate the input content or take the average input length. However, the text length can vary significantly across social media platforms, which may lead to missing important information and to an increase in the divergence between the inputs. To optimise the input length, we add the Input Length Optimiser component that handles this that iterates through different lengths in search of the optimal value.

\textbf{Source Encoder \& Target Encoder.} The two encoders for source and target platforms are based on Transformers which are trained multiple times while being incrementally adapted. First, using the labelled source dataset to train the source encoder. Second, the source encoder will fully or partially share parameters with the target encoder for initialisation. Finally, the source and target encoders will jointly perform adversarial adaptive training.

\textbf{Hidden State Selector.} The Hidden State Selector assesses the transferability of each layer in the Transformer to obtain the most transferable layers for pre-adversarial training and post-adversarial training.

\textbf{Discriminator \& Encoder Measurer.} These two components are combined to form an adversarial network to train the target encoder. The learning goal consists in reconstructing the target encoder to map the target input representation to the source input space, making it difficult for the discriminator to determine which platform the input comes from.

The discriminator consists of two fully-connected layers on top of the encoders. The two-layer feed forward network is designed with Rectified Linear Unit (ReLU) activation and 512 or 3072 hidden sizes for the first layer and Softmax activation for the output layer. We then adopt a supervised loss function from the ADDA framework \cite{Tzeng2017a}.


During the experiments, we observed that when there is a big platform data shift, gradient vanishing is common, such as in the case of transferring between Twitter and Wikipedia. To solve this issue, we add the Encoder Measurer component. Its learning goal is to get a similar hypothesis when the target encoder and the source encoder confront the same source datasets. As the loss function, we adopt the Kullback–Leibler divergence (KLD) metric \cite{Eguchi2006}. These losses (Discriminator loss and Encoder Measurer loss) are then joined to train the target encoder.


\textbf{Adaptive Batch Normalisation classifier.} The Adaptive Batch Normalisation (BN) classifier aims to reduce the distribution difference between the source and the target data by adjusting the dimensionality of input representations from source and target platforms. Similar to the discriminator structure, a two-layer feed forward network is designed by using ReLU activation. We adopt two methods to build the first layer. Reduction consists in reducing the hidden states to 512, while expansion consists in expanding it to 3072. For the output layer, we use Softmax as the activation function. Batch normalisation is added to standardise these inputs and reduce the generalisation error, so as to increase the generalisation ability of the classifier \cite{Li2019}.

\section{Experiments}

\subsection{Datasets}

We evaluate XP-CB on three widely-studied cyberbullying datasets\footnote{We restrict to cyberbullying datasets avoiding conflation with related phenomena such as hateful / toxic / abusive content.} from three social media platforms: Formspring \cite{Reynolds2011}, Twitter \cite{Waseem2016} and Wikipedia \cite{Wulczyn}. Where datasets provide finer-grained labels for types of cyberbullying, we collapse them into a cyberbullying label. The text length of each platform (see Table \ref{tab:datasets}) varies greatly from a maximum length of 38 to 2,846; so does the cyberbullying ratio ranging from 0.08 to 0.32.

\begin{table}[htb]
  \caption{Dataset statistics.}
  \label{tab:datasets}
  \begin{tabular}{cccc}
    \toprule
                        & Formspring  & Twitter     & Wikipedia  \\
    \midrule
    {\#}Posts           & 12,773      & 16,090      & 115,864    \\
    Max Length          & 1099        & 38          & 2832       \\
    Cyberbullying Ratio & 0.08        & 0.32        & 0.11       \\
    \bottomrule
  \end{tabular}
\end{table}

\subsection{Experiment Setup}
\label{ssec:exp-setup}

We set up experiments in line with previous work \cite{Agrawal2018,Dadvar2018}. We conduct six cross-platform configurations for our experiments by defining all six possible combinations of source-target dataset pairs. We focus on zero-shot settings, where the model doesn't see any labelled instances of the target platform. For a fair comparison with previous work, we adopt the same approach to mitigate the class imbalance by over-sampling the training data from the bullying class thrice.

\textbf{Transformer models.} We test XP-CB with two different Transformer models: BERT\_base (uncased) and RoBERTa\_base. We use the hyper-parameters recommended by \cite{Sun2019a}; Batch size: 16; Learning rate (Adam): 2e -5; Number of epochs: 4.

\textbf{Baseline models.} We compare our models with three competitive baseline models: (1) the cross-platform BiLSTM model with attention by \citet{Agrawal2018}, (2) BERT\_base (uncased) and (3) RoBERTa\_base.

\section{Results}

Table \ref{tab:out-result} shows the Macro-averaged F1 (Macro-F1) scores of all models under study, including the state-of-the-art model by \citet{Agrawal2018} (A\&A) and the baseline Transformer models, BERT and RoBERTa.

\begin{table}[ht]
\tabcolsep=0.06cm
{%
  \begin{tabular}{c||c|c|c||c|c}
    \toprule
    & \multicolumn{3}{|c||}{Baselines}
    & \multicolumn{2}{|c}{XP-CB}
    \\
    \midrule
    Source$\rightarrow$Target & A\&A & BERT & RoBERTa & \textbf{-BERT} & \textbf{-Roberta} \\
    \midrule
    \multicolumn{6}{c}{In-platform} \\
    \midrule
    \texttt{TW $\rightarrow$ TW}                    & 0.93  & 0.95  & 0.86  &  --            & --   \\
    \texttt{WP $\rightarrow$ WP}                    & 0.87  & 0.86  & 0.88  &  --            & --   \\
    \texttt{FS $\rightarrow$ FS}                    & 0.91  & 0.88  & 0.87  &  --            & --   \\
    \midrule
    Average                                         & 0.903 & 0.897 & 0.870 & -- & -- \\
    \midrule
    \multicolumn{6}{c}{Cross-platform} \\
    \midrule
    \texttt{FS} $\rightarrow$ \texttt{TW}           & 0.03  & 0.43  & 0.46  & \textbf{0.58}  & \textbf{0.61}  \\
    \texttt{WP} $\rightarrow$ \texttt{TW}           & 0.28  & 0.47  & 0.51  & \textbf{0.53}  & \textbf{0.56}  \\
    \hline
    \texttt{FS} $\rightarrow$ \texttt{WP}           & 0.35  & 0.74  & 0.78  & \textbf{0.81}  & \textbf{0.82}  \\
    \texttt{TW} $\rightarrow$ \texttt{WP}           & 0.10  & 0.54  & 0.56  & \textbf{0.60}  & \textbf{0.60}  \\
    \hline
    \texttt{TW} $\rightarrow$ \texttt{FS}           & 0.07  & 0.63  & 0.66  & \textbf{0.71}  & \textbf{0.71}  \\
    \texttt{WP} $\rightarrow$ \texttt{FS}           & 0.58  & 0.78  & 0.78  & \textbf{0.88}  & \textbf{0.86}  \\
   \midrule
   Average                    & 0.235 & 0.598 & 0.625 & \textbf{0.685} & \textbf{0.693} \\
 \bottomrule
 \end{tabular}%
 }
 \caption{In-platform and cross-platform classification results. \textbf{FS}: Formspring; \textbf{WP}: Wikipedia; \textbf{TW}: Twitter.}
 \label{tab:out-result}
\end{table}

\begin{figure*}[htb]
  \centering
  \includegraphics[width=\linewidth]{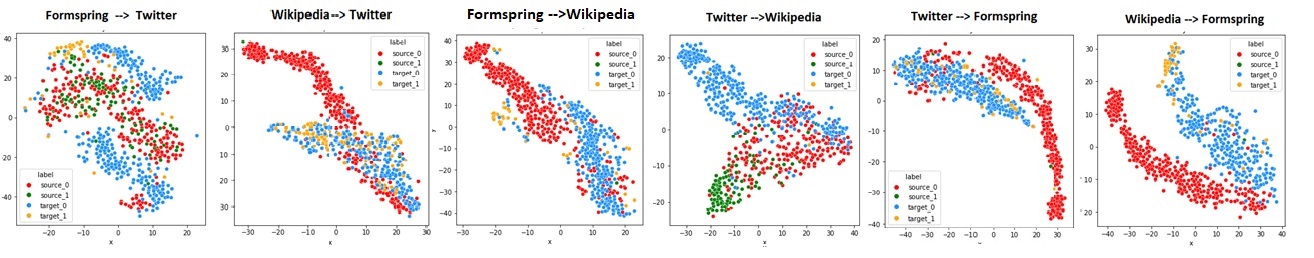}
  \caption{BERT t-SNE. Red:source negative;Green:source positive; Blue:target negative;  Yellow: target positive.}
  \label{fig:BERT}
  \end{figure*}

\begin{figure*}[htb]
  \centering
  \includegraphics[width=\linewidth]{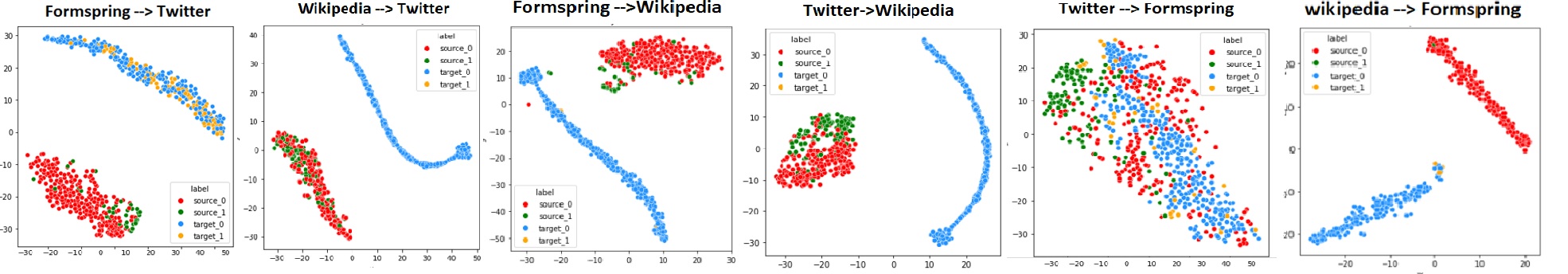}
  \caption{XP-CB t-SNE. Red:source negative;Green:source positive; Blue:target negative;  Yellow:target positive.}
  \label{fig:Framework}
  \end{figure*}

A look at the in-platform classification results shows the strong potential of the A\&A model, achieving slightly better average performance than Transformer models BERT and RoBERTa, despite this improvement not being consistent for all datasets. While BERT and RoBERTa models perform best for Twitter and Wikipedia respectively, it is the A\&A model that achieves the highest performance on Formspring. The high performance scores of these three models however drops substantially when we look at cross-platform experiments, with absolute performance drops of 67\% (A\&A), 30\% (BERT) and 25\% (RoBERTa) when we look at average performances. While RoBERTa demonstrates to be the best of the three models for cross-platform transfer, its performance is still seriously impacted.

We observe that the proposed XP-CB framework boosts this performance in all six cross-platform configurations, demonstrating its ability to further the cross-platform transferability of both BERT and RoBERTa encoders. XP-CB provides an absolute improvement of 9\% when we use BERT as the underlying encoder and an absolute improvement of 7\% when we use RoBERTa. These improvements are consistent across all six configurations, where the differences between XP-CB-BERT and XP-CB-RoBERTa are generally more marginal. We observe better overall performance when crossing between Formspring and Wikipedia in either direction, potentially due to the lengthier posts in both cases. Performance is lower for configurations involving Twitter, where the length is much shorter.

To assess the effectiveness of XP-CB in inferring representations that reconcile source and target platforms, we visualise the resulting embeddings for the six cross-platform configurations by using t\-SNE (t-Distributed Stochastic Neighbour Embedding) \cite{van2008visualizing}. Figure \ref{fig:BERT} displays the embeddings generated by the BERT\_base model.\footnote{We focus these visualisations on the BERT embeddings, rather than the RoBERTa embeddings, due to the limited space.} In some of the configurations (FS $\rightarrow$ TW, WP $\rightarrow$ TW and TW $\rightarrow$ WP), we can observe that the data points of different clusters in the source and target platforms are mixed together, which shows that a model trained on the source platform labelled data using only BERT is not enough for the target platform classification. Regarding the other three cross-platform configurations (FS $\rightarrow$ WP, TW $\rightarrow$ FS, WP $\rightarrow$ FS), the data points have begun to move closer to their own clusters. Although the boundaries of each group are not so clear, this shows that BERT has started to have some platform awareness.

Figure \ref{fig:Framework} shows the XP-CB embeddings. We can observe a clearer separation of classes with respect to the BERT embeddings, which demonstrates the increased platform-awareness of XP-CB. Especially on Wikipedia and Formspring, samples originating from different platforms are spatially separated. Along with the improved performance scores, visualisation of embeddings demonstrates the potential of XP-CB to bring the representations of source and target social media platforms closer to each other.
  
\section{Conclusion}
We have proposed XP-CB, a novel framework for cyberbullying detection in settings hitherto largely overlooked, i.e. across different social media platforms through zero-shot settings. Building on Transformer models BERT and RoBERTa, our framework couples its fine-tuning capacity with adversarial learning to enable cross-platform transfer. Through experiments on six cross-platform configurations, our study demonstrates the consistent effectiveness of XP-CB to outperform competitive baselines, including the state-of-the-art cross-platform cyberbullying detection model.


\bibliography{cyberbullyin}
\end{document}